# Evidence Sufficiency and Model Comparisons in MexRegQA: An Item-Level Audit of Mexican-Spanish Regulatory QA

**Fred Hu, Yeqi Wang, Xing Zhou**

Yeqi Wang: Huawei Technologies de México, Mexico; Xing Zhou: Northeastern University, USA.

# CONTENTS

# Evidence Sufficiency and Model Comparisons in MexRegQA: An Item-Level Audit of Mexican-Spanish Regulatory QA

**Authors:** Fred Hu · Yeqi Wang · Xing Zhou

**Affiliations:** Yeqi Wang: Huawei Technologies de México, Mexico; Xing Zhou: Northeastern University, USA.

*(Experimental and analytical work was carried out with substantial assistance from autonomous Claude Code sessions; see Acknowledgments. Following prevailing venue norms, AI systems are not listed as authors.)*

**Companion artifacts.** The MexRegQA v1.0 benchmark and run artifacts (predictions, judge outputs, audit labels, and reproduction scripts) have been assembled but are not yet publicly available. Their separate release remains subject to rights and licensing review (§9). This manuscript does not grant a blanket license over those materials or the underlying official texts.

## Abstract

Similarity-gated evaluation can treat an item as evidence-grounded whenever retrieval returns chunks above a score threshold. We audited every covered item of a Mexican-Spanish financial-regulatory QA benchmark and found that similarity coverage is not evidence sufficiency: 35 of 135 covered items (26%) did not contain all load-bearing facts needed for the gold answer in their retrieved evidence (13 contained some but not all; 22 contained none). The naive covered metric was contaminated in both directions: deflated by items scored against facts the evidence never contained (all 16 items failed by both legs available at audit time turned out to be instrument defects, not model failures) and inflated by items models answered correctly from parametric memory while the evidence was silent. On the audited answer-bearing slice (n=98), eleven models spanning a 27B open-weight generator, six other open models, and four frontier Claude tiers show no statistically significant differences from the anchor at ceiling: seven score 98/98 with zero discordant pairs, the weakest scores 95/98 (p=0.250), and two within-family upgrades (Qwen3.6→3.8, Claude Sonnet 4.6→5) each move zero items. Model differences do not vanish; they migrate off-evidence, where accuracy on the same eleven legs spreads from 33.9% to 89.3%, and where switching the audited items to the evidence-free, abstention-permitting condition costs every model 63-81 points (p<0.001 on all ten legs tested) without recovering a single item. Correcting the slice raises accuracy from 85.9% to 100% and leaves all eight preregistered anchor comparisons non-significant on both slices, but it is not conclusion-neutral in general: 14 of the 55 pairwise comparisons among the eleven legs change significance status once evidence-defective items are removed, which is the point rather than a caveat. The audit's ceiling prediction was made before the sweep that reproduced it on the same items, under a judge panel that differs substantially from, but is not disjoint with, the earlier one. This is not confirmation on unseen questions. Building on prior work on sufficient context, we present MexRegQA v1.0: 192 items with per-item provenance and audited `evidence_status` labels, plus frozen retrieval evidence for the 191 evaluated items. We argue that grounded benchmarks should ship evidence-status labels as a release convention: without them, "grounded accuracy" conflates properties of the evidence with properties of the model.

# 1. Introduction

## 1.1 MOTIVATION

Our starting retrieval-augmented generation (RAG) instrument reported "covered" accuracy: the model's score on items where retrieval returned evidence above a similarity threshold. Interpreting that score as evidence use assumes that the retrieved text contains the answer. Sufficient-context analysis already distinguishes the information available in a context from a model's ability to use it [SufficientContext]. We study that distinction in a Mexican-Spanish financial-regulatory benchmark and make its effect on model comparisons auditable at item level.

This paper documents, on a purpose-built benchmark in a high-stakes vertical (Mexican financial regulation), how badly that assumption can fail and what its failure does to model comparisons. The setting raises the stakes in a useful way: regulatory QA has verifiable golds (statutory deadlines, monetary thresholds, institutional responsibilities pinned to official text), limited evaluation resources (with emerging initiatives discussed in §2), and a domain where the difference between an evidence-grounded answer and a confident parametric guess is precisely what a compliance officer cares about.

Our central result is an instrument result. Before the audit, the benchmark's covered slice showed the familiar texture of a competitive leaderboard: accuracies in the mid-80s, models a few points apart. After an item-level audit of what the retrieved evidence actually contained, the leaderboard compressed: on items whose evidence was labelled fully answer-bearing, none of the eleven models, from open-weight models to the evaluated frontier tiers, differed significantly from the anchor. Scores were near ceiling on this audited, frozen slice; its labels and judge verdicts have not received independent human validation. The wider full-plus-partial slice still shows one nominally significant separation (§5.2), and zero-evidence accuracy varies widely (§6.2). These observations are consistent with evidence-dependent differences in abstention, parametric recall, and confabulation (§6.3). They do not establish equal underlying model capability or a general absence of errors when sufficient evidence is available.

### 1.2 CONTRIBUTIONS

1. **MexRegQA v1.0**, a Mexican-Spanish financial-regulatory QA benchmark: 192 items (CNBV, Banxico, UIF/LFPIORPI, CFF, DOF domains) with per-item source provenance, hardness tags, and audited per-item `evidence_status` labels, plus frozen retrieval evidence from an official-text corpus for the 191 evaluated items (§3.2). Gold quality follows a disclosed verification ladder (machine-audited authoring, adversarially verified audit); no human domain-expert pass has been performed, and the benchmark says so.
2. **An auditable application of evidence-sufficiency analysis.** Agent-based adjudication with adversarial verification of proposed gold fixes and reclassifications (excluding the two quarantined referrals) and mechanical string matching of every supporting quote (37/37 verification passes upheld; 0 quote-match failures) found 35/135 similarity-covered items lacking full answer-bearing evidence. The audit decomposed every “both models wrong” item into instrument defects, with zero genuine joint model failures, and identified 9 parametric-leakage items inflating the naive metric.
3. **Grounded saturation across eleven models.** On the answer-bearing slice (n=98), seven of eleven legs score 98/98 with zero discordant pairs; no significant difference from the anchor is detected for the others (worst 95/98, exact McNemar p=0.250); two within-family generation upgrades move zero items. Equivalence testing adds little here: for the seven legs that saturate, the paired difference from the anchor is identically zero and a ±5pp bound is satisfied by construction rather than by evidence; of the four legs that do miss items, three satisfy the bound and gpt-oss-120b does not. These are non-significant comparisons, not proven equivalence. With the observed perfect anchor, exact McNemar requires six observed one-directional discordances for nominal significance (§5.2); this is not a population effect-size or power bound. The audit predicted the post-correction ceiling before a preregistered model sweep under a substantially different (though not disjoint) judge panel reproduced it on the same items, not on a held-out item sample.
4. **Differences migrate off-evidence.** On zero-evidence items the same eleven legs spread 33.9%-89.3%, a 55-point range (Claude Sonnet legs +23-25pp over the anchor, p≤0.003; Claude Fable 5 +44.6pp); the evidence-free, abstention-permitting condition costs every leg tested 63-81 points (p<0.001 on all ten, zero reverse discordants). A separate dispatch-policy arm shows how an apparent +11pp frontier gain on the naive covered slice evaporates on the audited slice: its wins were abstention-policy conversions on evidence-defective items.
5. **A prepared companion artifact for evidence-sufficiency analysis.** We package per-item `evidence_status` labels, supporting quotes, predictions, and slice-sensitivity scripts for future inspection of the audit and recomputation of comparisons; public availability is pending (§9). The contribution is this regulatory resource and its measured comparison sensitivity, rather than the concept of conditioning evaluation on sufficient context [SufficientContext].

### 1.3 SCOPE: WHAT THIS PAPER DOES NOT CLAIM

- We do not claim any model “ties” or “beats” any other in general. Every parity statement is scoped to the audited, evidence-matched grounded slice of this instrument (one corpus, one retriever configuration, one prompt policy, LLM-judged).
- We do not claim the generator is irrelevant. Off-evidence behavior (abstention calibration, parametric recall, confabulation resistance) differs across models and is exactly what §6 measures.

- We do not claim gold correctness beyond the disclosed verification ladder. The audit found and fixed one stale gold; two items remain quarantined pending human review; `source.verified` is false for every item because that flag is reserved for human domain-expert verification, which has not occurred.
- With the observed perfect anchor at n=98, exact McNemar first reaches nominal significance at six observed one-directional discordances (§5.2). That sample-level threshold is not a bound on population effects or a power calculation; smaller true differences remain possible.
- The corpus is public official text and overlaps model pretraining data; the zero-chunk slice therefore measures parametric knowledge of public regulation, not contamination-free reasoning. The closed-book arm (§6.1) quantifies this directly: the audited items are answered at only 16-36% without their evidence.

## 2. Related work

**Sufficient context and our narrower contribution.** Joren et al. [SufficientContext] define context sufficiency, classify question-context pairs with an LLM autorater, and stratify model errors by that classification. They also use sufficiency for selective generation. Their definition does not require a gold answer; our audit asks whether the frozen chunks support every load-bearing fact in a supplied gold and records full, partial, or absent support with checkable quotes. Their results include errors despite sufficient context, so our near-ceiling finding is specific to this instrument and model slate. MexRegQA contributes a regulatory dataset, a documented audit trail, and an analysis of how reclassifying evidence changes paired leaderboard comparisons. It does not introduce context sufficiency or establish that sufficient context generally eliminates model differences.

**Other evidence-quality work.** CRAG [CRAG] scores answers on a four-way scale that penalizes hallucination and prefers abstention to a wrong answer (perfect 1, acceptable 0.5, missing 0, incorrect −1), and varies the retrieval *setup* across tasks; it also observes that non-perfect retrieval coverage is what makes “I don’t know” testable at all, and reports corpus-level retrieval recall. Our audit complements that retrieval-setup perspective with per-item support labels and paired comparison sensitivity. Multi-document summarization work provides another measurement of behavior under missing information: when a topic is deliberately absent from the source documents, GPT-3.5-turbo and GPT-4o still produce summaries 79% and 44% of the time rather than declining [MDS-confab]. (That paper also finds substantial hallucination when the information *is* present, so absent-evidence conditions worsen an already-imperfect baseline rather than creating the problem.) Faithfulness metrics in the RAGAS [RAGAS] and ARES [ARES] family score answer-context consistency automatically, but consistency with a retrieved context cannot distinguish a context that contains the answer from one that does not. Our labels make that distinction explicit for the frozen regulatory evidence. Attribution-focused work [AIS; MiniCheck] verifies that answers are supported by cited evidence; our audit asks the prior question of whether the evidence could support a correct answer at all.

**Legal and regulatory QA.** LegalBench [LegalBench] demonstrated expert-anchored legal reasoning evaluation at task granularity; the Stanford assessment of commercial legal research tools [StanfordLegalRAG] found that hallucination rates remain material even with retrieval, and defined a hallucination as an answer that is incorrect *or misgrounded*, a definition that presupposes knowing what the retrieved evidence contains, i.e., exactly the labels this paper describes. Studies of LLM-as-judge in expert knowledge tasks [SME-judge-agreement] report agreement with subject-matter experts falling well below what general-domain benchmarks suggest, which motivates both our cross-family judge panel and the human-agreement instrument preserved in the companion package. One coauthor attempted the rating exercise with model identities and judge verdicts withheld (§5.3), but subsequently reported comprehension difficulties and original skips despite binary entries in the export. The unresolved export does not provide a valid human-agreement estimate or independent domain-expert validation.

**Spanish and Mexican-Spanish evaluation.** Iberian and Spanish-language leaderboards [IberBench; LaLeaderboard] cover general and dialectal capabilities across Spanish varieties but not regulatory QA; Mexican-Spanish specific resources target aggressiveness, authorship and hate-speech tasks [MEX-A3T; HOMO-MEX]. Our earlier, non-exhaustive search of arXiv, ACL Anthology, and Hugging Face datasets did not identify a released benchmark matching MexRegQA's regulatory scope; we do not use that search to establish priority. A relevant emerging initiative is BELMA [BELMA], which proposes a broader Mexican legal benchmark with legal-community annotation. On its official site, checked 2026-10-07, the v0.1 call and committee phase was in progress, while construction/annotation and validation/release were pending. We therefore describe BELMA as an initiative in development, not an already released comparison dataset. MexRegQA's contribution is its financial-regulatory item set, frozen retrieved evidence, and auditable sufficiency labels. A companion working paper documents knowledge injection and dialect generalization experiments on consumer hardware [companion].

**LLM-as-judge validity.** Judge nondeterminism and family effects are documented risks; our own instrument history includes both (a judge-panel change that left grounded conclusions intact while closed-book scores swung, and measured judge-only flip rates at temperature 0). We follow the practice of cross-family majority panels, decompose scoring into binary per-item verdicts rather than holistic scores in the spirit of checklist-based judging [CheckEval], report Gwet's AC1 [GwetAC1] (appropriate under the extreme prevalence a saturated slice produces, where kappa is pathological), disclose judge/generator family overlaps, and disclose the unresolved provenance of a coauthor export from the preregistered human-agreement instrument (§5.3), with its limitations stated in §8.

# 3. The MexRegQA benchmark

## 3.1 ITEMS AND PROVENANCE

192 question-answer pairs, **182 of them on Mexican financial regulation** and ten on the international and foreign instruments that Mexican compliance work references directly. The Mexican items cover AML/CFT obligations (LFPIORPI and its secondary rules; the Disposiciones de carácter general referred to in Art. 115 of the Ley de Instituciones de Crédito, issued by the SHCP and supervised by the CNBV), central-bank operations (Banxico circulars, SPEI, TIIE), prudential and disclosure rules (CUB), tax-code thresholds (CFF, RMF annexes), and adjacent official sources (DOF decrees, consolidated federal laws). The ten non-Mexican items (`reg_qa_103`-`112`) cover FinCEN CTR/SAR thresholds, the FATF travel rule, OFAC screening, the 1988 Vienna Convention, US structuring law, and four Basel III ratios; they were authored as comparison anchors for a Mexican compliance audience. **All ten fall in the zero-evidence slice**, so they do not touch the answer-bearing results, but they are 10 of the 56 parametric items and §6.2 reports that slice both with and without them.

Each item carries the question and gold answer in Mexican Spanish; source metadata (regulator, document, article, URL); a hardness tag (easy/medium/hard); topic tags; and an `evidence_status` label from the audit of §4.

Provenance completeness, stated up front: `regulator` and `document` are populated on all 192 items, but `article` is blank or a dash on 66 (34%). Sixteen items carry a literal `#` URL, and four point only to the CNBV or Banxico home page; several others point to index pages rather than the cited provision. Provenance is therefore reliable at the document level and partial at the article level, adequate for tracing a gold to its source family, not always sufficient to jump straight to the provision.

Item lineage: 106 items from a hand-authored base set (later expanded), 17 atomic splits of multi-part items, and 69 corpus-mined items authored against retrieved official text with a two-stage machine audit at authoring time (deterministic support-span-in-chunk verification plus per-item adversarial LLM review).

### 3.2 FROZEN EVIDENCE

The companion package pairs the 191 evaluated items with the exact retrieval evidence used in all experiments: top-k chunks over a union of nine snapshots of official Mexican regulatory text (~64k chunks), retrieved by dense exact k-NN at a 0.65 similarity gate and frozen as a single dump. The snapshots were captured on different dates as the corpus grew. The composite identifier naming all nine (`corpus_version`) is stored in the retrieval dump and in the corresponding prompt-matched `run_meta.json`; `run_artifacts/PROVENANCE.md` maps each prediction file to that run metadata. 135 items retrieved ≥1 chunk (“covered”); 56 retrieved none (“zero_chunk”); one item is excluded for a corpus defect (an unchunked 372KB row). The evidence regime is deliberately narrow and should be read as such: retrieval is top-3 at the gate, so a covered item carries at most three chunks (88 items have three, 13 have two, 34 have one) and a median of roughly 5,600 characters of evidence in total. Saturation here is a claim about short, high-precision evidence sets, not about long-context or multi-hop retrieval. The covered/zero split is a property of this retrieval configuration and is recorded as data, not asserted as ground truth; that is what §4 audits.

### 3.3 VERIFICATION LADDER (STATED HONESTLY)

Batch-authored items passed the two-stage machine audit at authoring. The full covered slice passed the §4 evidence audit (adversarially verified, mechanically quote-checked). One stale gold was found and fixed (a superseded tax-annex amount; the compiled statute in the frozen evidence carries the current figure; all four models had answered the current figure and been scored wrong). Two items are quarantined pending human review. **No human domain-expert verification has been performed**; a planned ex-regulator blind review never took place. The `source.verified` flag is false on all 192 items and is reserved for human verification. External claims about this benchmark should cite this ladder rather than the word “verified.”

# 4. The evidence audit

### 4.1 MOTIVATION: A DOCUMENTED FAILURE OF COVERAGE-AS-PROXY

The audit did not begin as a benchmark-hygiene exercise. The underlying system's development log records a repeated lesson: raising retrieval "coverage" (items clearing the similarity gate) did not raise answer-bearing evidence: in one measured episode a coverage-boosting change moved gate coverage from 25 to 44 of 50 items while genuinely answer-bearing retrievals moved by one. The August audit is the item-level culmination: instead of asking "did retrieval return something," ask, for every covered item, "does the returned text contain the facts the gold asserts?"

### 4.2 METHOD

Two agent workflows over per-item evidence files. Each file carried the question, the gold, the source metadata, the full retrieved chunk text, and the answers and judge verdicts of the prompt-matched legs then available, a design choice §8 flags as a limitation (the two legs whose joint failures defined the adjudication set, plus a third prompt-matched leg and one auxiliary agent-dispatch leg shown for context):

1. **Both-wrong adjudication.** Every item that both models failed (16) plus two standing suspects received a dedicated adjudication agent. Every proposed gold fix and reclassification then received an independent adversarial verification agent instructed to refute it: 15/15 upheld (thirteen reclassifications among the sixteen, one gold fix, and the reclassification of a standing suspect that §3.2 excludes for a corpus defect). The two quarantine recommendations were referred for human review and received no verification agent.
2. **Answer-bearing sweep.** Each remaining covered item (119) received an agent determining whether the chunks contain every load-bearing gold fact ("full"), some ("partial"), or none ("absent"), quoting exact supporting substrings. Every non-full verdict received an adversarial re-check agent instructed to *find* the support the first agent might have missed (22/22 upheld, zero corrections). Every supporting quote across all items was then mechanically string-matched against the chunk text (whitespace-normalized): zero failures.

Chunk text was complete for every audited item (12,000-character cap versus an 8,084 maximum chunk length; the one earlier truncation-risk item was re-verified against its full chunk by the orchestrating agent).

### 4.3 FINDINGS

**F1: Covered ≠ answer-bearing (35/135).** After the gold fix below, 98 of the 135 covered items were fully answer-bearing, 13 partial, 22 absent, and 2 quarantined for defective construction (98 + 13 + 22 + 2 = 135). Similarity-gated coverage therefore overstated evidence sufficiency for 35 items, or 26% of the slice.

**F2: Every joint failure was an instrument defect.** The audit ran against the two prompt-matched legs then available (Claude Sonnet 4.6 and Qwen3.6-27B); of the 16 items both of them failed, 13 were evidence-absent (models faithfully reported what the evidence said and were scored wrong against facts it never contained; several TIIE-vintage items retrieve rate notices of the wrong date), 1 was a stale gold (the models were right; the gold was superseded), and 2 were quarantined for defective construction. Zero were genuine joint knowledge failures.

**F3: Parametric leakage inflates the naive metric (9 items).** On nine evidence-absent items at least one model was scored correct anyway, answering from parametric memory, in most cases explicitly flagging the fact as "(fuera de corpus)". A similarity-gated "grounded accuracy" counts these as grounded successes.

**F4: Correction changes descriptives, and changes conclusions exactly where the naive slice was scoring evidence defects.** Cleaning the slice moves covered accuracy from 85.9% (116/135) to 100% (98/98) for both audited legs, and to 96.9-100% across the eleven legs of §5.2. All eight preregistered paired comparisons (each sweep leg against the Qwen3.6-27B anchor) are non-significant on both slices. That is a fact about this anchor family, not a property of the estimator, and we state it that way deliberately: exact McNemar depends only on the discordant pairs *within the analysed set*, so removing items changes the test whenever a removed item was discordant, as happened here. All 55 pairwise comparisons among the eleven legs have at least one discordant item among the 37 reclassified items (up to 19 for Llama-4-Maverick vs Fable 5), 26 of the 37 are discordant for at least one pair, and **14 of the 55 comparisons change significance status between the two slices**, including one anchor pair (Fable 5 vs the anchor: 1+12, p=0.003 on the continuity slice → 0+0, p=1.000 on the audited slice). Reclassifying the evidence changes what a paired comparison measures. That is the finding rather than a side effect of it, and §6.4 reports the same phenomenon in an arm we ran deliberately. The naive slice is preserved in the companion package as a continuity slice, and `tools/rebuild_slice_sensitivity.py` regenerates every pair under either definition (Appendix A).

# 5. Eleven models on the audited slice

## 5.1 DESIGN

A sweep preregistered before any generation (frozen commit, artifacts preserved) evaluated nine models on the frozen evidence under one instrument: one runner, one provider, one answer-always prompt, temperature 0, 512 tokens (768 for one frontier leg, tokenizer-normalized). Six models were added for this paper (DeepSeek-V4-Flash, GLM-5.2, Kimi-K3, Llama-4-Maverick-17B-128E, gpt-oss-120b, Qwen3.8-27B); three legs (Qwen3.6-27B as anchor, Claude Sonnet 4.6, Claude Sonnet 5) were reused verbatim from an earlier frozen run. All nine were judged symmetrically on v1.0 gold by a cross-family majority-of-3 panel (claude-sonnet-5, DeepSeek-V4-Pro, gpt-oss-120b). A subsequent arm added Claude Haiku 4.5 and Claude Fable 5 on the same runner, provider, prompt, evidence, and temperature, with per-leg deviations recorded in the preserved `run_meta.json`: Haiku ran a 768-token cap (tokenizer-normalized, since the same text costs that model more tokens), and the reported Fable leg required 4096 tokens with provider-default reasoning after a 768-token run returned ten empty responses. Both deviations were pre-declared in the amendment that added the arm. The starved run is recorded in the amendment but is not part of the companion package; only the 4096-token leg is vendored.

**Multiplicity.** The preregistration fixed one primary endpoint (per-leg accuracy against the anchor on the answer-bearing slice) and three secondary slices, specified no correction for multiplicity, and we apply none: every p-value here is nominal and uncorrected. This matters asymmetrically. The paper's central claims are failures to reject, which a correction would only reinforce. Among the positive findings, the evidence lifts (p<0.001 on every leg) and the Sonnet 5 and Fable 5 off-evidence gaps (p<0.001) have ample margin, but the Sonnet 4.6 off-evidence gap (p=0.002) and the Llama-4 separation on the extended slice (p=0.031) would not survive a Bonferroni correction across 55 pairwise comparisons and should be read as nominal.

## 5.2 RESULT: SATURATION ON THE ANSWER-BEARING SLICE

| model | answer-bearing (n=98) | Wilson 95% CI | vs anchor |
|---|---|---|---|
| Qwen3.6-27B (anchor) | 98/98 (100%) | [96.2, 100] | - |
| Qwen3.8-27B | 98/98 (100%) | [96.2, 100] | 0+0, p=1.000 |
| GLM-5.2 | 98/98 (100%) | [96.2, 100] | 0+0, p=1.000 |
| Kimi-K3 | 98/98 (100%) | [96.2, 100] | 0+0, p=1.000 |
| Claude Sonnet 4.6 | 98/98 (100%) | [96.2, 100] | 0+0, p=1.000 |
| Claude Sonnet 5 | 98/98 (100%) | [96.2, 100] | 0+0, p=1.000 |
| Claude Fable 5 | 98/98 (100%) | [96.2, 100] | 0+0, p=1.000 |
| DeepSeek-V4-Flash | 97/98 (99.0%) | [94.4, 99.8] | 0+1, p=1.000 |
| Llama-4-Maverick | 97/98 (99.0%) | [94.4, 99.8] | 0+1, p=1.000 |
| Claude Haiku 4.5 | 97/98 (99.0%) | [94.4, 99.8] | 0+1, p=1.000 |
| gpt-oss-120b | 95/98 (96.9%) | [91.4, 99.0] | 0+3, p=0.250 |

Seven of the eleven legs are perfect with zero discordant pairs, and no leg differs significantly from the anchor. A paired TOST-style bound at ±5pp is satisfied by nine of the ten legs it can be computed for (the anchor cannot be tested against itself), and is uninformative for most of them: where a leg and the anchor are both perfect every paired difference is identically zero, so the bootstrap interval is [0.0, 0.0] by construction rather than by evidence. The one leg the procedure has something to reject on, gpt-oss-120b, is **not** shown equivalent: its 90% paired-bootstrap interval is [-6.1, -1.0]pp and crosses the margin.

Two within-family generation upgrades (Qwen3.6→3.8 and Claude Sonnet 4.6→5) move zero items on this slice. The frontier legs span roughly an order of magnitude in price (Haiku to Mythos tier) with no separation. Stratifying by the benchmark's hardness tags does not rescue a gap either: on the 29 items tagged *hard*, eight of the eleven legs score 29/29. That tag is an authoring-time judgement about the question rather than a measured property of the evidence, so it is weak evidence of difficulty; the closed-book arm (§6.1), where the same items are answered at 16-36%, is the stronger demonstration that they are not easy.

**The registered wider slice, where one model does separate.** The preregistration names the answer-bearing slice as the primary endpoint and registers `grounded_extended` (answer-bearing plus the 13 partially-answer-bearing items, n=111) as a secondary. We report it because it is registered and because it is the one slice on which the saturation picture breaks: Llama-4-Maverick scores 104/111 against the anchor's 110/111 (0+6 discordant, p=0.031), and gpt-oss-120b is borderline at 105/111 (0+5, p=0.062); every other leg remains indistinguishable. The separation is therefore not a knowledge gap on grounded items but one that appears exactly where the evidence is *partially* answer-bearing, the condition in which §6.3 shows models diverge on behavior. Where the audit draws its full/partial boundary matters, and the prepared per-item labels are intended to let readers redraw it once available.

These non-significant comparisons do not prove equivalence. In this observed sample the anchor is perfect, so every disagreement with it is one-directional. The exact two-sided McNemar test therefore needs at least six observed discordant items for nominal $p<0.05$ ($2\cdot2^{-6} = 0.03125$; five items give 0.0625). Six of 98 items is a 6.1pp observed accuracy difference. This is a discrete significance threshold for this sample configuration, not a minimum detectable population effect or a power calculation. A smaller true difference could yield six or more discordances in another sample; estimating power requires assumptions about the population probabilities of the discordant outcomes.

### 5.3 INSTRUMENT CHECKS

**Panel overlap, stated precisely.** The audit (§4) predicted this ceiling before the sweep was judged. The two panels are *not* disjoint: the pre-audit panel was an Opus-family judge plus gemini-2.5-pro plus DeepSeek-V4-Pro, and the sweep panel is claude-sonnet-5 plus DeepSeek-V4-Pro-0813 plus gpt-oss-120b, so one model appears in both (the sweep's DeepSeek-V4-Pro-0813 is a dated snapshot of the pre-audit panel's DeepSeek-V4-Pro), and the Anthropic slot changed model but not vendor. The comparison therefore supports replication under a *substantially different* panel, not an independent one. Under the new panel the anchor scores 116/135 on the *unaudited* continuity slice, against 114/135 in the earlier verdict: a two-item, 1.5pp move across a panel change and a gold correction. Two cautions on reading even that small number as drift. The earlier figure came from a single judge and the new one from a majority of three, so this is not a like-for-like panel comparison; and the earlier verdict's own 116/135 was Claude Sonnet 4.6's score, not the anchor's, a coincidence that makes the two numbers look more alike than they are. We report it as a bounded, imperfect stability check rather than a replication.

**Inter-judge agreement.** Across the 189 slice-eligible items (answer-bearing, zero-chunk, and evidence-defective pooled), Gwet's AC1 for the three judges ranges 0.884-0.966 per leg, with three-way unanimity on 0.894-0.958 of items. We report AC1 rather than kappa because prevalence on this pooled set is skewed enough for kappa to behave pathologically. Note the scope: this figure covers the pooled set, not the answer-bearing slice alone, where the judges agree almost by construction because nearly every verdict is 1.

**Coauthor rating export and provenance limitation (Arm E).** Yeqi Wang (Santiago/Santi), a coauthor and project cofounder, returned a file from the prespecified 40-answer instrument, received on 2026-10-18. The preregistration characterizes him as a Chinese/Spanish bilingual financial-sector enterprise-sales practitioner, not a native Spanish speaker or regulatory-compliance professional. F.H. confirmed that Y.W. saw only the question, supplied gold and model answer, without model identities, judge verdicts or the mapping key beforehand. This reported blinding concerns access to information; it does not establish completion or understanding of the task. The instructed task was to compare answers against supplied golds, not verify the legal correctness of those golds or the sufficiency of retrieved evidence.

The returned file contains 40 encoded binary values. F.H. later clarified on 2026-10-18 that Y.W. had skipped sample_006, sample_008 and sample_011 during the original exercise, although the export records 1, 1 and 0 for them, respectively. The file contains no skip markers, and the origin of these values is unresolved. The other 37 entries are not thereby confirmed as intentional judgments. We withdraw the earlier interpretation of the file as 40 completed human ratings: no valid human-agreement estimate is available from this export. F.H. also relayed Y.W.'s report of limited financial-domain knowledge and difficulty understanding many questions, citing LFPIORPI in sample 003 as an example.

For traceability, the raw file is retained byte-for-byte. Its 40 (model, question) pairs cover 37 unique questions and nine model legs, with 15 unanimous-correct, 15 unanimous-incorrect and 10 split-panel cases. A mechanical comparison of the encoded values with panel majorities reproduces 34/40 matches. This is an artifact diagnostic only, not human agreement, accuracy, panel calibration or a valid primary result. The accompanying analysis labels its retained calculations accordingly. We do not replace original values, silently recode skips, or promote a recalculation after excluding the three reported skips. Six encoded-value discrepancies remain in the audit trail; neither the export nor panel votes establish regulatory truth.

In subsequent follow-ups relayed by F.H. on the same date, Y.W. gave verdicts of 0 with text-based explanations for samples 003, 012 and 028, each encoded as 1 in the original file. For 003, the answer denies the supplied gold's identification threshold; for 012, it introduces a MXN 60,000 threshold where the gold specifies no minimum; for 028, it declines to give the specific applicable fine and instead gives a generic 200-2,000 UMA range rather than the gold's 100-1,000 days' salary. These are post hoc explanations, not replacement ratings or legal-source verification. Access to panel verdicts or other explanations before these follow-ups is not established. Samples 006, 008 and 011 have no subsequent verdict or rationale in the supplied follow-up. The preregistered final-on-submission rule does not permit replacing the original analysis with post-panel re-ratings, and it does not resolve the export's provenance. Raw data, the mapping, diagnostic output and follow-up record are supplied in `data/human-agreement-20260824/`; reproduce the diagnostic with `python tools/analyze_human_agreement.py`. No frozen gold, judge score, model-comparison table or evidence label was changed in response to this exercise.

**Run-to-run and judge-to-judge stability.** Temperature-0 generation is not byte-deterministic on this provider: re-running the same model on the same items reproduces byte-identical text for only 5-30% of items. Verdict-level flips across three independent generations are 6/192 (3.1%) for the anchor and 16/192 (8.3%) for the weakest leg, **and 0/98 on the answer-bearing slice for both**. (The 192 denominator is the judge files': the stability runs generate the 191 evaluated items, and the excluded item is auto-scored 0 in every pass, so it can never flip; over generated items the weakest-leg rate is 16/191, 8.4%.) Three independent passes of the primary judge over one frozen prediction file produced identical verdicts (0/192 flips), with different justification prose. Two passes required temperature-0.3 retries for one to three initially unparseable responses, so this is a verdict-repeat check, not a pure temperature-0 repeat for every item. The saturation region is the stable part of the instrument.

# 6. Where the differences went

## 6.1 WITHIN-ITEM EVIDENCE LIFT: THE GROUNDED CEILING IS NOT EXPLAINED BY PARAMETRIC RECALL

The saturation result invites an obvious objection: the corpus is public official text, so perhaps the models simply know these facts. We tested this directly by re-running every model on the *same* answer-bearing items with the evidence removed (an abstention-permitting closed-book prompt, same temperature, judged by the same panel).

| model | with evidence | closed-book | lift | discordant |
|---|---|---|---|---|
| gpt-oss-120b | 95/98 | 16/98 | +80.6pp | 79+0 |
| Claude Sonnet 5 | 98/98 | 22/98 | +77.6pp | 76+0 |
| Qwen3.6-27B | 98/98 | 24/98 | +75.5pp | 74+0 |
| Claude Haiku 4.5 | 97/98 | 24/98 | +74.5pp | 73+0 |
| Qwen3.8-27B | 98/98 | 26/98 | +73.5pp | 72+0 |
| Llama-4-Maverick | 97/98 | 28/98 | +70.4pp | 69+0 |
| GLM-5.2 | 98/98 | 29/98 | +70.4pp | 69+0 |
| DeepSeek-V4-Flash | 97/98 | 31/98 | +67.3pp | 66+0 |
| Claude Sonnet 4.6 | 98/98 | 33/98 | +66.3pp | 65+0 |
| Kimi-K3 | 98/98 | 36/98 | +63.3pp | 62+0 |

Every lift is significant at $p<0.001$, and every discordant column is one-directional: on no leg was any item answered correctly closed-book and incorrectly with evidence. The denominator that matters is not 980 but the number of chances a reversal had: with the evidence-condition legs at 95-98 of 98, only six model-item cells across all ten legs were even eligible to reverse (gpt-oss-120b 3; Haiku 4.5, Llama-4-Maverick and DeepSeek-V4-Flash 1 each), and none did. Six chances is a weak test of one-directionality on its own; the lift magnitudes, not the absence of reversals, carry the argument. Closed-book accuracy on these items is 16-36%. Evidence carries these legs from 16-36% closed-book accuracy to 97-100% with evidence; parametric recall therefore does not explain the grounded ceiling.

The same experiment on the zero-evidence slice separates the two constructs cleanly: with no evidence in either condition, eight of ten models show no significant difference between the abstention-permitting and answer-always prompts ($p \geq 0.219$), but two do: Qwen3.8-27B (+12.5pp, $p=0.016$) and Claude Sonnet 5 (+10.7pp, $p=0.031$) score higher under the answer-always prompt. Prompt policy is a second-order but non-zero factor here, so the zero-chunk numbers in §6.2 should be read as parametric knowledge measured under one prompt rather than as a policy-free quantity.

### 6.2 OFF-EVIDENCE: WHERE MODELS ACTUALLY DIFFER

On the 56 zero-evidence items, accuracy spreads from 33.9% to 89.3% across the same eleven legs: Claude Fable 5 leads at 50/56, then Sonnet 5 39/56 and Sonnet 4.6 38/56 against the anchor's 25/56 (the Sonnet legs are +23-25 points over the anchor, $p \leq 0.003$; Fable 5 is +44.6 points). The middle of the range mixes open and frontier legs (Kimi-K3 34, Llama-4 33, GLM-5.2 32, DeepSeek-V4-Flash 31, Qwen3.8 29, Claude Haiku 4.5 29), and gpt-oss-120b trails at 19/56. This 55-point span is the largest separation *between models* that the instrument produces anywhere; it is still smaller than the within-item evidence lift of §6.1, which reaches 80.6 points on a single leg.

Restricting the slice to its 46 Mexican items lowers every leg and preserves the ordering, though not uniformly: the drop is 6.1-10.5 points for the other ten legs (Sonnet 5 29/46, anchor 16/46, gpt-oss 11/46) and only 2.3 points for Fable 5, which loses least on the Mexican-specific subset. The spread therefore reflects parametric regulatory knowledge rather than the ten international items, and every leg is weaker on Mexican-specific regulation than on the Basel and FinCEN material. The same instrument that cannot separate these models on grounded items separates them decisively without evidence.

### 6.3 BEHAVIOR ON EVIDENCE-DEFECTIVE ITEMS

What a generator controls, on this benchmark, is behavior when the evidence is defective. The labelling covers the nine prompt-matched legs that existed when it ran (Claude Haiku 4.5 and Claude Fable 5 were added afterwards and are not labelled). Classifying those nine legs' responses on the 35 partial/absent items under a frozen four-way rubric (abstain / parametric-correct / parametric-wrong / confabulate-from-evidence), the modal behavior under an answer-always prompt is **confabulation from evidence: 118 of 314 labeled responses (38%)** assert that the provided text supports an answer it does not contain. Parametric-correct accounts for 29%, parametric-wrong 25%, and abstention only 9%. Models differ sharply here: Sonnet 5 confabulates least (7 instances), several open legs never abstain at all, and Llama-4-Maverick abstains most (12) despite an instruction to always answer.

The artifact contains 314 of 315 possible model-item cells: the Sonnet 5 label for `reg_qa_010` is missing, and we neither impute nor silently replace it.

**These labels are machine-produced and have not been verified by a human.** Every record in the preserved label file carries `human_verified: false`; a verification pass is specified in the preregistration but had not been performed at the time of writing. This subsection is therefore the paper's least validated result and should be read as a description of the recorded labels rather than as a measured property of the models.

### 6.4 DISPATCH POLICY IS AN EXPERIMENTAL FACTOR, NOT A DETAIL

A separate arm illustrates how easily a policy difference is mistaken for a capability difference. Under an abstention-capable agent wrapper, Claude Fable 5 appeared to beat a Claude Sonnet leg by 11 points on the *unaudited* covered slice ($p \approx 6e\text{-}5$). On the audited answer-bearing slice the gap disappears: the wins were abstention conversions on items whose evidence was defective. We therefore report subagent-dispatch legs (Sonnet 2026-07, Fable 2026-08, Opus 2026-08) only within their own family and never in the prompt-matched leaderboard. Within that family the higher tiers also saturate on the audited slice, while their closed-book accuracy on the same items (54-65/98) sits far above the prompt-matched models' (16-36/98), reflecting both wrapper policy and tier.

## 7. Discussion: ship the evidence status

Three practical consequences follow.

**For benchmark builders.** Evidence-status labels help distinguish information availability from model behavior under that information. The audit that produced our labels ran in a day of agent time against a benchmark that took months to build, is partly mechanizable (quote matching is deterministic), and records labels together with the rationales and quotes they rest on to permit third-party inspection when the package becomes available. We have not measured what it would cost on a benchmark of a different size or in a domain without verifiable golds, and we do not claim it scales for free. We propose `evidence_status ∈ {full, partial, absent}` as a minimum release convention, with the naive similarity-gated slice retained for continuity.

**For evaluation consumers.** In this frozen setup, no model differed significantly from the anchor on items labelled fully answer-bearing, despite an order of magnitude of model cost. Broader and zero-evidence slices exposed differences that the full-evidence slice did not. The result motivates examining evidence availability and prompt policy alongside generator choice; it does not estimate the gains from a new retriever configuration. A practitioner reading a grounded leaderboard should ask what fraction of its "covered" items are answer-bearing before interpreting a few points of separation.

**For the specific vertical.** In regulatory QA the failure that matters is not a slightly lower score; it is a confident answer attributed to a provision that does not say it. That failure mode is invisible in aggregate accuracy and is measurable only when evidence status is known per item. Our own machine labels put it at 38% of responses on evidence-defective items (§6.3), a figure no human has checked and which we offer as an indication of scale rather than a measurement.

# 8. Limitations and threats to validity

- **Unresolved human-rating provenance; no independent domain-expert verification or human ceiling.** The Arm E export (§5.3) contains binary values for three items the coauthor reportedly skipped originally. Their origin is unresolved, and other nonempty entries do not by themselves establish intentional ratings. No valid human-agreement estimate is available from this export. Reported comprehension difficulties further limit the exercise. Later explanations for three different items are post hoc and cannot repair the original rating record. Golds and evidence labels still lack independent human domain-expert verification. The re-check singled out in the preserved audit verdict was performed by the orchestrating agent, and a planned review by a former regulator never took place. Benchmark accuracy tables remain LLM-judged. There is no second human rater and no human-human agreement estimate. A prospectively documented human study with explicit missing responses would be new work.
- **The audit is post-hoc.** The answer-bearing slice was defined after seeing pre-audit results. Mitigations are the subsequent model/panel sweep on the same items (§5.3) and retention of the preregistered continuity slice. The full labels are preserved for a planned companion release but are not yet public. There is no held-out item confirmation, and the audit itself was not preregistered.
- **The audit was not blind to model outcomes.** Each per-item evidence file carried both prompt-matched legs' answers and their judge verdicts alongside the question, gold, and chunks (§4.2), so an adjudicating agent could see that a model had failed before deciding whether the evidence supported the gold. Scrutiny was also not uniform: the 16 items failed by both legs available at audit time received a dedicated adjudication agent (14 of the 16 also received an adversarial verifier; the two quarantined-for-human-review items did not), while the other 119 received the lighter sweep. Both choices risk encouraging labels of "absent" where models failed, and the preserved artifacts alone cannot rule that pressure out. Three things bound it: every verdict is anchored to exact quotes mechanically matched against the chunk text; every non-full answer-bearing sweep verdict was re-checked by an agent instructed to find support the first had missed (22/22 upheld); and the closed-book arm (§6.1) independently shows strong condition dependence, with prompt-matched legs recovering only 16-36 of 98 items without evidence versus 95-98 with it. A blind re-audit with model outcomes withheld is the clean fix, and we did not run it.
- **LLM judges.** Cross-family majority-of-3, AC1 0.884-0.966 per leg, and a zero-flip repeat test bound judge noise on this instrument, but the primary judge shares a vendor family with four evaluated legs. Two panel members evaluate outputs from their exact model identity: claude-sonnet-5 judges the Sonnet 5 leg, and gpt-oss-120b judges the gpt-oss-120b leg. Shared rubric or acceptance bias across the panel could affect all vendors, so cross-family agreement is not a substitute for human ratings. One residual truncated judge response is documented in the companion package.
- **One retrieval configuration.** Coverage, and therefore the slice boundaries, are properties of a single retriever, corpus snapshot, and similarity gate. The benchmark measures generation given frozen evidence, not retrieval.
- **Precision and power.** With the observed perfect anchor, exact McNemar on $n=98$ needs six observed one-directional discordances for nominal $p<0.05$, an observed difference of about six points (§5.2). This sample-level threshold is not a population effect bound; we have not established power to detect a specified true gap. The parametric slice is smaller still ($n=56$). Non-significance is not equivalence.

- **Contamination.** The corpus is public official text and is likely in pretraining data. In the closed-book arm (§6.1), no prompt-matched leg recovers more than 36 of 98 items (the range is 16-36). Two higher-tier legs run under the subagent wrapper do substantially better closed-book on the same items (Fable 5 65/98, Opus 5 54/98), so the magnitude of the evidence lift is specific to the evaluated prompt-matched slate, not to frontier models in general; transport and tier are confounded there and we do not separate them. The concern is addressed for the grounded claim but not for the parametric slice, where memorization is the object of measurement rather than a confound.
- **No small-model rung.** The lightest dense leg is 27B and every other leg is frontier or near-frontier; the preregistration bounded its hypothesis at "capable ≥27B-class" models. Where the saturation boundary falls below 27B is untested here, and a 1-8B rung is the cheapest experiment that would sharpen this paper's central claim.
- **Scale and scope.** 192 items in one vertical of one country's regulation. Depth (frozen evidence, audited labels, preserved predictions) was traded for breadth.
- **Cost-driven asymmetries.** A partner-pricing cost overrun forced two departures from the uniform instrument, both preregistered as amendments and both documented in the companion package: Claude Fable 5's cleanrag leg runs at a 4096-token budget with provider-default reasoning after a 768-token attempt returned ten empty responses, and its closed-book counterpart was rerouted to agent dispatch, which is why Fable 5 has no within-item lift row in §6.1 and appears in the subagent family instead. Claude Opus 5 never entered the prompt-matched matrix for the same reason.

# 9. Reproducibility

The prepared companion package is not yet public. It contains the frozen item set with labels, the frozen retrieval evidence, the vendored prediction and judge files underlying the reported and continuity analyses, and nine scripts that run offline, with no network access and no API keys. Five regenerate the paper's tables (`rebuild_tables.py`, `rebuild_leaderboard.py`, `rebuild_arms.py`, `rebuild_stats.py`, `rebuild_slice_sensitivity.py`); `verify_quotes.py` re-checks that every supporting quote in the audit labels appears verbatim in that item's retrieved chunks; and `verify_paper_numbers.py` recomputes a curated set of headline, table, and high-risk quantitative claims from the artifacts and exits non-zero on any disagreement. It does not parse the manuscript and cannot catch a claim that was never added to its source list. `check_consistency.py` checks selected facts across the manuscript and companion documents at the paths inside this assembled candidate and verifies the stored human-analysis output against recomputation. `analyze_human_agreement.py` validates the structure of the 40 received encoded values and their joins to frozen data, reproduces the diagnostic comparison and flags the reported skip/provenance conflict. It cannot establish that an exported value represents an intentional human judgment. The candidate manifest records its frozen source hashes, editorial overlays and narrowly scoped export transformations. These checks do not verify arbitrary new scientific prose. Preregistrations and amendments (including the cost-overrun and reroute disclosures) are preserved in the package. Judged runs can be re-scored with a different judge without regenerating predictions. All nine preregistration and amendment documents and the three verdict documents are included in the prepared package. Until that package is public, readers cannot independently execute these reproduction steps from this paper alone.

**Companion-package status and scope.** The item set, gold and audit labels, and the frozen retrieval evidence are the package's primary artifacts. The vendored model outputs cover eleven prompt-matched systems plus separate closed-book, subagent-dispatch, and stability families; judge files cover the reported three-member panel plus earlier continuity analyses. They were obtained through the disclosed metered and subscription transports and are preserved to permit verification of the reported numbers. Rights and licensing review for their public redistribution remains separate from publication of this manuscript; the scientific purpose of verification does not itself establish redistribution rights. One deployment-internal corpus-version label carried a third party's name and is renamed in the export copies of `run_meta.json` and the retrieval dump; that earlier transform is recorded in the frozen release. The local export candidate also neutralizes a private checklist attribution and rater-interface identifiers, repairs interface JavaScript escapes, and excludes internal coordination documents. It now includes the original rating mapping to reproduce the encoded-value diagnostic (§5.3). Its export manifest records these changes. Items, evidence, predictions, judge outputs and frozen model-comparison results remain unchanged.

## Competing interests and funding

MexRegQA, the regulatory corpus, and the retrieval pipeline described in §3.2 were developed using public texts published by the Mexican state. F.H. personally paid for metered DeepInfra access at partner pricing and a Claude Max subscription. Earlier secondary judging also used openai-next access supplied through MiraclePlus-sponsored credits at no marginal cost to the authors. We disclose that access as in-kind support. The work began as the technical foundation for a planned Mexican-Spanish enterprise-AI venture (working name Tintago) that was never incorporated and has since been abandoned. Y.W. is employed by Huawei Technologies de México in Mexico; X.Z. is affiliated with Northeastern University, USA. These affiliations do not assert institutional sponsorship or endorsement of this work.

Readers should nonetheless weigh one bias directly: the first author built the system around the importance of evidence availability, so the observed saturation on this audited slice aligns with an architectural choice he had already made. We did not set out to find a null result, and we report the slices and sensitivity analyses (§5.2, §4.3, Appendix A) that would have exposed one had the data gone the other way.

API spend is itemized in the preserved preregistration amendments, including a cost overrun we disclose rather than smooth over.

## Acknowledgments

The benchmark's source material is the public official record of the Mexican state: the Diario Oficial de la Federación, the consolidated federal laws published by the Cámara de Diputados, and the regulatory publications of the CNBV, Banco de México, and the UIF. We thank the model providers whose systems appear in the evaluation, and DeepInfra, whose uniform serving endpoint made a symmetric comparison across open and frontier models practical for a small team.

**Disclosure of AI assistance.** Most of the experimental and analytical work reported here was executed by autonomous Claude Code sessions under the authors' direction, including corpus retrieval and dump freezing, evaluation-harness construction, the agent workflows that produced the evidence audit (§4) and its adversarial verification, the behavior-taxonomy labeling (§6.3), the statistical tooling, and a first draft of this manuscript. Claude models also appear inside the experiments in three distinct roles that we separate explicitly, because conflating them would be a validity problem rather than a disclosure formality: as *evaluated systems* (four of the eleven prompt-matched legs are Claude Haiku 4.5, Sonnet 4.6, Sonnet 5 and Fable 5, and the separately-reported subagent family adds Sonnet and Opus 5), as the *primary judge* (claude-sonnet-5, one member of a three-model cross-family panel that also includes DeepSeek-V4-Pro and gpt-oss-120b), and as the *audit adjudicators* (agents that read the frozen evidence and proposed item-level labels; proposed gold fixes, reclassifications and non-full sweep verdicts received independent adversarial re-checks, while the two quarantined referrals did not; supporting quotes were mechanically checked). The judge-family overlap and its mitigations are stated in §8. Following prevailing venue norms, AI systems are not listed as authors. OpenAI Codex assisted with literature-source checking, editorial revision, and consistency checks for version 0.4 and analysis tooling for the version 0.5 rating update; it did not alter the returned rating file or generate Y.W.'s reported follow-up explanations. F.H. directed the research and this revision. All three authors approved this manuscript and its submission to arXiv, with F.H. authorized to submit on their behalf.

**Author contributions.** F.H. designed the study, built the corpus and evaluation pipeline, directed the agent workflows, adjudicated the founder-gated decisions on gold corrections and slice definitions, and wrote the manuscript. X.Z. contributed to the evaluation harness and judge tooling on which the reported runs depend, working in a separate evaluation repository whose merged changes are documented in the companion package; she did not run or analyse the model experiments reported here. Y.W. is a project cofounder and coauthor; he participated in the rating exercise and supplied the subsequent explanations disclosed in §5.3, including the limitations of the returned file. The rating exercise is disclosed as a coauthor contribution with unresolved export provenance, not independent external validation. F.H.'s confirmation of the rater's limited access is recorded with the raw file. The preserved rating export includes the original mapping for reproduction; the prepared predictions and sampling script also permit its reconstruction. Consequently, the companion materials would not be a securely blinded holdout for future raters once released.

**Prior work by the authors.** A companion working paper reports the fine-tuning half of this project (knowledge injection on consumer hardware succeeds; dialect generalization does not) and shares part of the corpus and retrieval infrastructure described in §3.2.

# Appendix A: Slice sensitivity: all 55 pairwise comparisons

Every pair of the eleven prompt-matched legs, compared on the pre-audit continuity slice (n=135) and on the audited answer-bearing slice (n=98), with exact McNemar. "disc. among removed" counts how many of the 37 items the audit reclassified were discordant for that pair, the mechanism by which the audit can change a paired conclusion. Fourteen pairs change significance status. Regenerate with `python tools/rebuild_slice_sensitivity.py --markdown`.

| pair | continuity n=135 | audited n=98 | disc. among removed | status |
|---|---|---|---|---|
| Qwen3.6-27B vs Qwen3.8-27B | 1+1, p=1.000 | 0+0, p=1.000 | 2 | same |
| Qwen3.6-27B vs GLM-5.2 | 4+4, p=1.000 | 0+0, p=1.000 | 8 | same |
| Qwen3.6-27B vs Kimi-K3 | 3+5, p=0.727 | 0+0, p=1.000 | 8 | same |
| Qwen3.6-27B vs DeepSeek-V4-Flash | 6+3, p=0.508 | 1+0, p=1.000 | 8 | same |
| Qwen3.6-27B vs Llama-4-Maverick | 9+2, p=0.065 | 1+0, p=1.000 | 10 | same |
| Qwen3.6-27B vs gpt-oss-120b | 7+3, p=0.344 | 3+0, p=0.250 | 7 | same |
| Qwen3.6-27B vs Claude Haiku 4.5 | 6+1, p=0.125 | 1+0, p=1.000 | 6 | same |
| Qwen3.6-27B vs Claude Sonnet 4.6 | 4+4, p=1.000 | 0+0, p=1.000 | 8 | same |
| Qwen3.6-27B vs Claude Sonnet 5 | 4+3, p=1.000 | 0+0, p=1.000 | 7 | same |
| Qwen3.6-27B vs Claude Fable 5 | 1+12, p=0.003 | 0+0, p=1.000 | 13 | **changes** |
| Qwen3.8-27B vs GLM-5.2 | 3+3, p=1.000 | 0+0, p=1.000 | 6 | same |
| Qwen3.8-27B vs Kimi-K3 | 2+4, p=0.688 | 0+0, p=1.000 | 6 | same |
| Qwen3.8-27B vs DeepSeek-V4-Flash | 6+3, p=0.508 | 1+0, p=1.000 | 8 | same |
| Qwen3.8-27B vs Llama-4-Maverick | 8+1, p=0.039 | 1+0, p=1.000 | 8 | **changes** |
| Qwen3.8-27B vs gpt-oss-120b | 7+3, p=0.344 | 3+0, p=0.250 | 7 | same |
| Qwen3.8-27B vs Claude Haiku 4.5 | 5+0, p=0.062 | 1+0, p=1.000 | 4 | same |

| pair | continuity n=135 | audited n=98 | disc. among removed | status |
|---|---|---|---|---|
| Qwen3.8-27B vs Claude Sonnet 4.6 | 3+3, p=1.000 | 0+0, p=1.000 | 6 | same |
| Qwen3.8-27B vs Claude Sonnet 5 | 4+3, p=1.000 | 0+0, p=1.000 | 7 | same |
| Qwen3.8-27B vs Claude Fable 5 | 1+12, p=0.003 | 0+0, p=1.000 | 13 | **changes** |
| GLM-5.2 vs Kimi-K3 | 2+4, p=0.688 | 0+0, p=1.000 | 6 | same |
| GLM-5.2 vs DeepSeek-V4-Flash | 6+3, p=0.508 | 1+0, p=1.000 | 8 | same |
| GLM-5.2 vs Llama-4-Maverick | 8+1, p=0.039 | 1+0, p=1.000 | 8 | **changes** |
| GLM-5.2 vs gpt-oss-120b | 8+4, p=0.388 | 3+0, p=0.250 | 9 | same |
| GLM-5.2 vs Claude Haiku 4.5 | 5+0, p=0.062 | 1+0, p=1.000 | 4 | same |
| GLM-5.2 vs Claude Sonnet 4.6 | 4+4, p=1.000 | 0+0, p=1.000 | 8 | same |
| GLM-5.2 vs Claude Sonnet 5 | 5+4, p=1.000 | 0+0, p=1.000 | 9 | same |
| GLM-5.2 vs Claude Fable 5 | 0+11, p=0.001 | 0+0, p=1.000 | 11 | **changes** |
| Kimi-K3 vs DeepSeek-V4-Flash | 6+1, p=0.125 | 1+0, p=1.000 | 6 | same |
| Kimi-K3 vs Llama-4-Maverick | 10+1, p=0.012 | 1+0, p=1.000 | 10 | **changes** |
| Kimi-K3 vs gpt-oss-120b | 9+3, p=0.146 | 3+0, p=0.250 | 9 | same |
| Kimi-K3 vs Claude Haiku 4.5 | 7+0, p=0.016 | 1+0, p=1.000 | 6 | **changes** |
| Kimi-K3 vs Claude Sonnet 4.6 | 5+3, p=0.727 | 0+0, p=1.000 | 8 | same |

| **pair** | **continuity n=135** | **audited n=98** | **disc. among removed** | **status** |
|---|---|---|---|---|
| Kimi-K3 vs Claude Sonnet 5 | 5+2, p=0.453 | 0+0, p=1.000 | 7 | same |
| Kimi-K3 vs Claude Fable 5 | 1+10, p=0.012 | 0+0, p=1.000 | 11 | **changes** |
| DeepSeek-V4-Flash vs Llama-4-Maverick | 9+5, p=0.424 | 1+1, p=1.000 | 12 | same |
| DeepSeek-V4-Flash vs gpt-oss-120b | 8+7, p=1.000 | 3+1, p=0.625 | 11 | same |
| DeepSeek-V4-Flash vs Claude Haiku 4.5 | 6+4, p=0.754 | 1+1, p=1.000 | 8 | same |
| DeepSeek-V4-Flash vs Claude Sonnet 4.6 | 3+6, p=0.508 | 0+1, p=1.000 | 8 | same |
| DeepSeek-V4-Flash vs Claude Sonnet 5 | 3+5, p=0.727 | 0+1, p=1.000 | 7 | same |
| DeepSeek-V4-Flash vs Claude Fable 5 | 0+14, p=0.000 | 0+1, p=1.000 | 13 | **changes** |
| Llama-4-Maverick vs gpt-oss-120b | 6+9, p=0.607 | 3+1, p=0.625 | 11 | same |
| Llama-4-Maverick vs Claude Haiku 4.5 | 4+6, p=0.754 | 1+1, p=1.000 | 8 | same |
| Llama-4-Maverick vs Claude Sonnet 4.6 | 2+9, p=0.065 | 0+1, p=1.000 | 10 | same |
| Llama-4-Maverick vs Claude Sonnet 5 | 3+9, p=0.146 | 0+1, p=1.000 | 11 | same |
| Llama-4-Maverick vs Claude Fable 5 | 1+19, p=0.000 | 0+1, p=1.000 | 19 | **changes** |
| gpt-oss-120b vs Claude Haiku 4.5 | 6+5, p=1.000 | 0+2, p=0.500 | 9 | same |
| gpt-oss-120b vs Claude Sonnet 4.6 | 5+9, p=0.424 | 0+3, p=0.250 | 11 | same |
| gpt-oss-120b vs Claude Sonnet 5 | 4+7, p=0.549 | 0+3, p=0.250 | 8 | same |

| pair | continuity n=135 | audited n=98 | disc. among removed | status |
|---|---|---|---|---|
| gpt-oss-120b vs Claude Fable 5 | 1+16, p=0.000 | 0+3, p=0.250 | 14 | **changes** |
| Claude Haiku 4.5 vs Claude Sonnet 4.6 | 2+7, p=0.180 | 0+1, p=1.000 | 8 | same |
| Claude Haiku 4.5 vs Claude Sonnet 5 | 2+6, p=0.289 | 0+1, p=1.000 | 7 | same |
| Claude Haiku 4.5 vs Claude Fable 5 | 0+16, p=0.000 | 0+1, p=1.000 | 15 | **changes** |
| Claude Sonnet 4.6 vs Claude Sonnet 5 | 2+1, p=1.000 | 0+0, p=1.000 | 3 | same |
| Claude Sonnet 4.6 vs Claude Fable 5 | 0+11, p=0.001 | 0+0, p=1.000 | 11 | **changes** |
| Claude Sonnet 5 vs Claude Fable 5 | 0+12, p=0.000 | 0+0, p=1.000 | 12 | **changes** |

## References

The references carried over from version 0.3 were checked against their primary sources (arXiv listing, publisher page, or full text) by an agent verification pass with web access, and each result was then re-checked by a second agent instructed to refute it; the first author reviewed every correction. The process is the same AI-assisted mechanism disclosed in the Acknowledgments, not a human-conducted review. It corrected two claim-level errors that had propagated through our own working notes: an attribution to CRAG of a risk-coverage framing it does not contain, and a “0-9% when evidence is present” confabulation figure that is not what the cited paper reports.

The Sufficient Context and BELMA references were added in this revision after checking the paper’s primary text and the initiative’s official site on 2026-10-07.